\documentclass[preprint,10pt,number]{elsarticle}

\usepackage{graphicx}
\usepackage{tikz}
\usetikzlibrary{shapes.geometric, arrows, arrows.meta, positioning, shadows}
\usepackage{pifont}
\usepackage{pgfplots}
\pgfplotsset{compat=1.18}
\usepackage{amsmath, amssymb, amsfonts}
\usepackage{url}
\usepackage{hyperref}
\usepackage{listings}
\usepackage{xcolor}
\usepackage{geometry}
\usepackage{algorithm}
\usepackage{algpseudocode}
\usepackage{caption}
\usepackage{longtable}
\usepackage{booktabs}
\usepackage{multirow}
\usepackage{natbib}

\newcommand{\cmark}{\ding{51}} 
\newcommand{\xmark}{\ding{55}} 

\tikzstyle{startstop} = [rectangle, rounded corners, minimum width=3cm, minimum height=1cm, text centered, draw=black, fill=blue!20]
\tikzstyle{process} = [rectangle, minimum width=3cm, minimum height=1cm, text centered, draw=black, fill=green!20]
\tikzstyle{decision} = [rectangle, minimum width=1cm, minimum height=1cm, text centered, draw=black, fill=yellow!20]
\tikzstyle{block} = [rectangle, draw, fill=green!20, minimum width=3.5cm, minimum height=1cm, text centered]
\tikzstyle{feature} = [rectangle, draw, fill=blue!20, minimum width=3.5cm, minimum height=1cm, text centered]
\tikzstyle{gate} = [circle, draw, fill=yellow!30, minimum size=0.7cm, text centered]
\tikzstyle{output} = [rectangle, draw, fill=blue!20, minimum width=3.5cm, minimum height=1cm, text centered]
\tikzstyle{epoch} = [rectangle, draw, fill=blue!10, minimum width=1.5cm, minimum height=1cm, text centered]
\tikzstyle{module} = [rectangle, draw, fill=green!20, minimum width=1cm, minimum height=0.7cm, text centered]
\tikzstyle{growth} = [red, thick, dashed]
\tikzstyle{arrow} = [thick,->,>=stealth]

\journal{}

\begin{document}

\begin{frontmatter}

\title{Adaptive Modular Retrieval-Augmented Framework for Cross-Domain Image Classification}

\author[1]{Debojyoti Ghosh \footnote{Corresponding Author: debojyoti07.dg@gmail.com}}
 \author[2]{Soumya K Ghosh}
\author[1]{Adrijit Goswami}

\affiliation[1]{organization={Department of Mathematics, Indian Institute of Technology Kharagpur},
                addressline={Kharagpur}, 
                postcode={721302}, 
                state={West Bengal}, 
                country={India}}
\affiliation[2]{organization={Department of Computer Science and Engineering, Indian Institute of Technology Kharagpur},
                addressline={Kharagpur}, 
                postcode={721302}, 
                state={West Bengal}, 
                country={India}}

\begin{abstract}
Effective image classification in real-world scenarios requires models capable of capturing both local features and global context, while adapting to the evolving complexity of the input. This paper presents a hybrid deep learning system named AMR-Net (Adaptive Modular Retrieval Network) for adaptable image classification in environmental and industrial contexts. The proposed model integrates ResNet-based local feature extraction and Vision Transformer-based global representation learning with an FAISS-based memory retrieval module for contextual enhancement. Furthermore, AMR-Net has an adaptive modular extension mechanism to dynamically enhance network capacity when validation performance plateaus during training. The proposed framework was evaluated using a ten-class waste classification dataset and the Kolektor Surface Defect Dataset 2 (KolektorSDD2), achieving classification accuracies of 97.15\% and 99.80\%, respectively. Experimental results demonstrate that AMR-Net consistently outperforms mainstream convolutional and transformer-based approaches while maintaining robust cross-domain generalization. Component-wise ablation studies show that global attention (ViT), retrieval-based contextual learning (FAISS), and adaptive modular growth each provide complementary and partially overlapping contributions to overall performance, with global attention contributing the largest individual gain. The proposed architecture is a scalable and flexible solution for image classification challenges characterized by data unpredictability, class imbalance, and dynamic feature distributions.
\end{abstract}


\end{frontmatter}

\section{Introduction}    

The accelerating pace of urbanization, rising consumption, and industrial activity has led to a substantial increase in global waste production, intensifying the challenges of effective waste management. Accurate and efficient waste classification plays a central role in enhancing recycling processes, reducing environmental impact, and promoting sustainability. Conventional manual sorting approaches remain widely used despite being labor-intensive, inconsistent, and prone to error and health risk \cite{1}. These limitations have motivated growing interest in automated solutions, particularly those based on deep learning.

Convolutional Neural Networks (CNNs) have shown remarkable proficiency in image-based waste classification, enabling automatic identification of waste categories without human supervision \cite{2}. By learning rich hierarchical features from large volumes of image data, these networks can distinguish among recyclables, hazardous materials, and organic waste with high accuracy \cite{3}. Nonetheless, several challenges persist. Many waste categories share overlapping visual characteristics, which complicates class separation, and the scarcity of diverse, large, well-annotated datasets constrains effective model training \cite{4}. Waste images also exhibit significant variability in color, texture, and size \cite{5}; transfer learning, data augmentation, and synthetic sample generation are commonly employed to mitigate this variability and improve robustness \cite{6}.

The heterogeneous nature of waste materials adds further complexity: categories such as plastics, metals, paper, and organic substances each have distinct visual signatures \cite{7}, and mixed or overlapping waste increases labeling difficulty. Hybrid architectures that combine multiple deep learning paradigms, or that fuse multiple sensor modalities, have been explored to leverage complementary strengths, for instance combining CNNs with recurrent networks to capture spatial and temporal patterns \cite{9}. The computational demands of deep models are also a practical concern for resource-constrained deployment settings such as smart bins or embedded systems \cite{10}; model compression strategies including quantization, pruning, and lightweight backbones such as MobileNet have been proposed to reduce cost while preserving performance \cite{11}, and hybrid pipelines that combine deep features with classical classifiers such as support vector machines have also been investigated \cite{1}.  

The convergence of deep learning with Internet of Things (IoT) technologies is gaining traction in intelligent waste management. Smart bins integrated with sensors and embedded inference algorithms can categorize waste at the point of disposal, enabling automated sorting and real-time feedback \cite{13}. Scaling such systems for widespread deployment, however, requires models that can adapt to novel waste categories and evolving temporal waste patterns; transfer learning has emerged as a practical mechanism for efficiently fine-tuning pretrained networks on new data with minimal retraining effort \cite{14}.

Unlike static deep learning models with fixed representational capacity, real-world classification systems must adapt to evolving data distributions, class imbalance, and increasing task complexity. Existing approaches typically rely on predefined architectures that cannot dynamically adjust their capacity during training. To address this limitation, we propose \textbf{AMR-Net}, an adaptive modular retrieval-augmented framework that integrates multi-scale feature extraction, retrieval-augmented memory, and dynamic capacity expansion. AMR-Net pairs the local feature extraction strength of a ResNet-50 backbone with the global reasoning ability of a Vision Transformer (ViT), and further incorporates a FAISS-based similarity retrieval module that acts as a memory system, retrieving related examples from the feature space to support context-aware decision-making. A key feature of AMR-Net is its adaptive design: rather than committing to a fixed architecture, the network includes expandable, learnable modular blocks that grow during training only when validation performance stagnates, allowing the model to scale its capacity according to task complexity.

A key motivation for combining these three components lies in their mutual reinforcement. A static, fixed-capacity network produces a feature manifold that drifts during training, which can progressively degrade the quality of nearest-neighbor retrieval from the FAISS memory bank. By allowing the network to grow only when validation improvement stagnates, the adaptive expansion mechanism helps stabilize the feature space, so that retrieved neighbors remain more semantically relevant throughout training. Conversely, higher-quality retrieved features provide a richer contextual signal to the modular blocks, which can help the network escape validation plateaus more effectively. This bidirectional relationship between adaptive capacity and retrieval-augmented memory motivates the joint design pursued in this work and is examined empirically in Section~\ref{RD}.

The subsequent sections of this work are organized as follows. Section~\ref{LS} provides a survey of relevant literature. Section~\ref{Model} details the AMR-Net architecture, including the hybrid integration of ResNet-50 and ViT, the adaptive modular design, and a dedicated theoretical motivation subsection. Section~\ref{sec:experimental-setup} describes the experimental setup, including dataset preprocessing, training configuration, and hyperparameter sensitivity analysis. Section~\ref{RD} presents an in-depth analysis of the experimental results, including ablation studies, component interaction analysis, and statistical significance tests. Section~\ref{App} presents an industrial application evaluation on the KolektorSDD2 dataset. Finally, Section~\ref{Cn} concludes the paper and outlines directions for future work.

\section{Literature Survey} \label{LS}

Recent years have shown substantial progress in applying deep learning to waste classification, spanning advanced convolutional architectures, hybrid learning strategies, efficiency-oriented model design, and multi-source data fusion. These developments coincide with rapid advances in industrial automation and smart environmental technologies, increasing demand for classification systems that are both adaptable and reliable.

\subsection{Deep CNN-Based Models}
Deep convolutional neural networks (CNNs) have become the cornerstone of many waste classification systems. Prior work includes the development of deep vision architectures to automate recycling workflows~\cite{1}. Data augmentation and transfer learning were applied to boost the performance of CNNs in domain-specific datasets, such as decoration waste~\cite{2}. A two-stage deep learning pipeline was introduced to enhance robustness in real-world applications~\cite{3}, while feature fusion strategies were adopted to improve human-robot interaction efficiency in smart recycling~\cite{4}. Several studies integrated pre-trained CNN backbones such as VGGNet~\cite{5}, or lightweight backbones like ShuffleNet v2 for mobile deployments~\cite{6}. Real-time performance was a focus of the MRS-YOLO model~\cite{7}, and SwinConvNeXt~\cite{madhavi2025swinconvnext} fused transformer and CNN architectures for accurate and efficient garbage image classification.

\subsection{Hybrid and Transfer Learning Models}
Several studies have explored the use of hybrid approaches and transfer learning methods to enhance waste classification. For instance, optimized convolutional neural networks have been developed to achieve both speed and accuracy, which are essential for real-time recycling applications~\cite{9}. Other research has combined convolutional neural networks with deep reinforcement learning techniques to improve the efficiency of sorting recyclable materials~\cite{10}. To tackle the challenges posed by occluded objects, attention mechanisms based on depth-wise separable convolutions have been introduced to boost the performance of CNNs~\cite{11}. In complex scenarios, integrating CNNs with autoencoders has shown to improve classification outcomes by extracting richer feature representations~\cite{1}. Reinforcement learning has also been utilized to enable models to adapt dynamically as new waste data becomes available~\cite{13}. Moreover, transfer learning paired with lightweight neural architectures has been applied to develop mobile-friendly waste classification systems~\cite{14}.

\subsection{Optimization and Lightweight Architectures}
Efforts to reduce model complexity without compromising accuracy are prominent in the literature. Several works apply depth-wise separable convolutions and pruning techniques for real-time classification on embedded devices~\cite{17}. Reinforcement learning-based optimization methods have also been proposed for dynamically adjusting classification strategies~\cite{18}. Other studies introduced edge computing architectures that combine local inference with cloud-based updates for high-throughput waste sorting~\cite{19}. Models like optimized CNN variants have been developed to minimize false positives and improve performance under noisy conditions~\cite{21}. \cite{dong2025improved} proposed an improved lightweight deep learning architecture for household waste detection, achieving efficient and accurate waste classification while reducing computational complexity, making it suitable for real-time smart waste management applications.

\subsection{Emerging Trends in Hazard and Industrial Waste Management}
Beyond traditional waste sorting, recent studies have expanded to hazard classification and intelligent industrial systems. For instance, a deep learning framework incorporating multifractal analysis was proposed for hazard severity assessment~\cite{n3}, while a Fourier-enhanced gray model combined with CNNs improved classification in dynamic environments~\cite{n2}. High-dimensional classification has benefited from QPSO-based fuzzy clustering~\cite{n1}, and CNNs have been used to classify and quantify recycled aggregates in real time~\cite{n6}. Furthermore, federated learning has been used to improve traceability and real-time performance in smart city waste management systems~\cite{n4}.

\subsection{Domain-Inspired Modular Architectures}
Recent domain-specific models illustrate a growing trend toward modularity and attention mechanisms for efficient classification. \cite{nezerka2024machine} demonstrated that handcrafted feature extraction can outperform traditional CNNs for construction and demolition waste. \cite{sun2025gd} proposed GD-YOLO, which improves YOLOv10n with slicing-based attention for high-speed edge deployment.

Numerous studies have highlighted key challenges and outlined prospective directions for advancing deep learning approaches in waste classification. \cite{31} introduced an approach that combines CNNs with deep reinforcement learning to improve classification performance in dynamic environments. Similarly, \cite{32} proposed a hybrid architecture that combines CNNs and RNNs, capturing both temporal and spatial information to enhance classification accuracy. In \cite{34}, a combination of CNNs and random forest classifiers was employed within smart recycling frameworks to improve sorting effectiveness. \cite{LONG2024108516} developed a robust plastic waste classification framework that integrates wavelet transform multi-resolution analysis with convolutional neural networks to enhance feature extraction and classification accuracy. \cite{JIANG2026106031} proposed a class-relevant patch embedding selection framework that identifies discriminative image patches to enhance feature representation and improve few-shot image classification performance. \cite{WU2025105647} introduced a Deep Hybrid Manifold Network with joint metric learning, which captures manifold structures and similarity relationships among image sets to achieve robust image set classification.

\cite{ghosh2025enhanced} demonstrated the effectiveness of FAISS-based similarity retrieval combined with deep convolutional feature extraction for waste classification. Their ResNet-152 and attention-enhanced framework achieved strong category discrimination but remained limited by a static architecture with fixed capacity and less adaptability to new domains. \cite{Feng2024} proposed a statistical adaptive modeling approach (SA-Det) with category statistics adaptive and distribution adaptive modules to improve kitchen waste detection in complex scenes. \cite{Tasic2025} developed convolutional neural networks optimized with a modified crayfish optimization algorithm, enhanced by AdaBoost and XGBoost, to achieve high-accuracy waste classification. \cite{GHOSH2026114602} proposed a semantic-augmented Vision Transformer with cross-modal fusion, achieving state-of-the-art performance in waste classification by integrating visual features with semantic attribute embeddings.

AMR-Net addresses these limitations through a hybrid and dynamically adaptive design that integrates localized feature extraction (ResNet-50) with global semantic representation (Vision Transformer). Inspired by biological adaptivity, the model incorporates a modular learning mechanism that allows the network to expand by adding new trainable blocks during training. This adaptive growth enables capacity scaling, enhancing flexibility and mitigating the rigidity of conventional deep networks. Evaluation on both waste classification and industrial surface defect detection (KolektorSDD2 dataset) demonstrates the model's robustness and cross-domain generalizability, providing a scalable and domain-agnostic framework for complex and evolving visual classification tasks.

A comparative analysis of recent state-of-the-art methods (see Table~\ref{tab:techniques-comparison}) illustrates that while prior work has explored various combinations of convolutional backbones, transformers, attention mechanisms, and multimodal inputs, few have simultaneously integrated adaptive growth and memory retrieval modules in a unified framework.

\begin{table*}[h!]
\centering
\caption{Comparison of architectures and strategies in recent waste classification studies}
\renewcommand{\arraystretch}{1.2}
\begin{tabular}{p{2.2cm} c c c c}
\toprule
\textbf{Model} & \textbf{CNN} & \textbf{Transformer} & \textbf{Adaptive} & \textbf{FAISS} \\
\midrule
\cite{7} & \cmark & \xmark & \xmark & \xmark \\
\cite{10} & \cmark & \xmark & \xmark & \xmark \\
\cite{madhavi2025swinconvnext} & \cmark & \cmark & \xmark & \xmark \\
\cite{sun2025gd} & \cmark & \xmark & \xmark & \xmark \\
\cite{ghosh2025enhanced} & \cmark & \xmark & \xmark & \cmark\\
\textbf{AMR-Net} & \cmark & \cmark & \cmark & \cmark \\
\bottomrule
\end{tabular}
\label{tab:techniques-comparison}
\end{table*}

The absence of adaptive growth in prior hybrid models carries a concrete practical cost. When a fixed-capacity CNN-Transformer encounters class-imbalanced or long-tailed data, it must either be overparameterized from initialization (risking overfitting on minority classes) or underparameterized (causing underfitting on visually complex categories). Neither option is satisfactory for datasets with the kind of imbalance present in the waste classification task, where the \textit{clothes} class contains 5,325 images while \textit{trash} contains only 697. Similarly, FAISS-based retrieval in a static architecture suffers from feature drift: as the backbone updates during training, previously stored embeddings become misaligned with the current feature space, reducing retrieval precision. AMR-Net addresses both limitations simultaneously. The adaptive growth mechanism allocates representational capacity in response to training dynamics, while the incremental FAISS update strategy ensures the memory bank reflects the current state of the feature extractor at every epoch.

\subsection{Research Gap and Motivation}

Despite substantial progress in deep learning for image classification, current models still face challenges in domain generalization, scalability, and adaptability to dynamic data environments. Many existing approaches rely on fixed-capacity architectures that struggle to transfer across datasets and lack mechanisms for incorporating memory or incremental learning. These limitations reduce their effectiveness in real-world scenarios characterized by class imbalance, evolving distributions, and environmental variability.

To address these challenges, we propose an adaptive modular framework that integrates a ResNet-50 convolutional backbone for local feature extraction, a Vision Transformer (ViT) for global contextual modeling, and a FAISS-based similarity retrieval module for memory-augmented representation learning. A key component of the proposed approach is a dynamic capacity expansion strategy that appends modular blocks when validation performance plateaus during training. This mechanism enables the network to scale its representational capacity in response to task complexity while maintaining efficient learning behavior. As a result, the framework is well suited for real-world waste classification tasks that exhibit long-tailed distributions, concept drift, and heterogeneous visual characteristics.

\subsubsection{Contributions}

In this paper, we introduce an adaptive modular retrieval-augmented classification framework designed to address persistent challenges in waste classification, including class imbalance, noisy labels, and evolving data distributions. The proposed approach aims to provide an accurate, robust, and scalable solution for real-world deployment in waste management systems.

The main contributions of this work are summarized as follows:

\begin{itemize}

\item \textbf{Adaptive Hybrid CNN--Transformer Framework with Memory Retrieval:}
We propose a unified architecture that combines a ResNet-50 backbone for local feature extraction, a Vision Transformer for global contextual modeling, and a FAISS-based similarity retrieval module that provides memory-augmented feature enhancement during classification.

\item \textbf{Dynamic Modular Capacity Expansion Mechanism:}
To overcome the limitations of fixed-capacity networks, we introduce an adaptive growth strategy that dynamically appends modular blocks when validation performance plateaus, enabling the model to increase representational capacity in response to training dynamics.

\item \textbf{Retrieval-Augmented Feature Fusion Strategy:}
A multi-branch fusion mechanism integrates convolutional features, transformer embeddings, and retrieved memory representations into a unified feature space, improving robustness in class-imbalanced and visually ambiguous scenarios.

\item \textbf{Cross-Domain Validation on Environmental and Industrial Tasks:}
The proposed framework is evaluated on both a large-scale waste classification dataset and the KolektorSDD2 industrial surface defect dataset, demonstrating strong performance and generalization across heterogeneous application domains.

\item \textbf{Comprehensive Experimental and Ablation Analysis:}
Extensive experiments, including component-wise ablation studies, a component interaction analysis, a qualitative failure-case analysis, and statistical significance tests, quantify the contributions of the transformer module, retrieval memory, and adaptive modular expansion.

\end{itemize}

Experimental results demonstrate that the proposed framework improves classification accuracy, precision, recall, and F1-score compared to standard baseline models. The adaptability and scalability of the approach make it suitable for deployment in evolving environments where data characteristics may change over time. By combining convolutional features, transformer-based context modeling, and memory-augmented retrieval within an adaptive architecture, AMR-Net offers a practical solution for automated waste sorting systems and other real-world classification settings. Figure~\ref{fig:tikz_flowchart_faiss_color} illustrates the workflow of the proposed classification model.

\begin{figure}[h!]
\centering
\begin{tikzpicture}[node distance=1.2cm, every node/.style={font=\small, align=center}]

\node[draw, rectangle, fill=blue!20, minimum width=3cm, minimum height=0.8cm] (input) 
{Input Images $x \in \mathbb{R}^{B \times 3 \times H \times W}$};

\node[draw, rectangle, right=of input, fill=red!20, minimum width=3cm, minimum height=0.8cm] (resnet) 
{ResNet-50\\(Extract ResNet features $f_\text{resnet}, 3840$)};

\node[draw, rectangle, below=of resnet, yshift=0.6cm, fill=green!20, minimum width=3cm, minimum height=0.8cm] (faiss) 
{FAISS Memory Bank\\Query $f_\text{resnet}$, retrieve $k$ neighbors\\Compute $f_\text{faiss} = \frac{1}{k}\sum f_i$\\Then store $f_\text{resnet}$};

\node[draw, rectangle, below=of input, yshift=-0.6cm, fill=yellow!20, minimum width=3cm, minimum height=0.8cm] (vit) 
{Vision Transformer\\Extract CLS token $f_\text{vit}, 768$};

\node[draw, rectangle, below=of faiss, fill=orange!30, minimum width=4cm, minimum height=1cm] (fusion) 
{Feature Fusion Layer\\$[f_\text{resnet}, f_\text{vit}, f_\text{faiss}] \rightarrow f_\text{fused}, 4096$};

\node[draw, rectangle, below=of fusion, fill=purple!20, minimum width=4cm, minimum height=1cm] (modular) 
{Adaptive Modular Blocks};

\node[draw, rectangle, below=of modular, fill=cyan!20, minimum width=3cm, minimum height=0.8cm] (classifier) 
{Classification Head};

\draw[->, thick] (input) -- (resnet);
\draw[->, thick] (input.south) -- (vit);
\draw[->, thick] (resnet) -- (faiss);
\draw[->, thick] (resnet.east) -- ++(1.8,0) |- (fusion.east);
\draw[->, thick] (vit.south)  |- (fusion.west);
\draw[->, thick] (faiss) -- (fusion);
\draw[->, thick] (fusion) -- (modular);
\draw[->, thick] (modular) -- (classifier);

\end{tikzpicture}
\caption{Integrated workflow showing retrieval-before-storage FAISS process, feature fusion, adaptive modular blocks, and final classification.}
\label{fig:tikz_flowchart_faiss_color}
\end{figure}

\section{Model Architecture}\label{Model}

\subsection{Pretraining and Fine-Tuning Strategy}
\label{subsec:pretraining}

The proposed framework leverages transfer learning to exploit rich visual representations learned from large-scale datasets. The convolutional backbone (ResNet-50) was initialized with weights pre-trained on the ImageNet-1K dataset, while the transformer backbone (ViT-Base) was initialized using weights pre-trained on ImageNet-21K.

To reduce computational cost and prevent overfitting on the relatively smaller target datasets, the backbone networks were frozen during training. Specifically, the convolutional layers of ResNet-50 and the transformer encoder layers of ViT-Base were kept fixed, and only the feature fusion module, adaptive modular blocks, routing gates, and classification head were trained.

Freezing the backbone layers preserves general-purpose visual features while allowing the newly introduced modules to specialize for the target tasks. This strategy also stabilizes training and reduces the number of trainable parameters, which is particularly important given the high capacity of the combined architecture.

The final fully connected layers of both backbones were replaced with task-specific modules to accommodate the number of waste categories. No partial fine-tuning was performed in the reported experiments, however, the framework allows selective unfreezing of deeper layers for domain-specific adaptation in future work.

\subsection{Feature Representation and Extraction}

AMR-Net employs a tri-branch feature extraction and contextual retrieval framework that integrates hierarchical convolutional representations, global transformer embeddings, and memory-enhanced similarity retrieval using a FAISS index. Let the input image be represented as

\[
x \in \mathbb{R}^{H \times W \times C}.
\]

\subsubsection{Local Feature Extraction via ResNet-50}

Hierarchical visual patterns are extracted using a pre-trained ResNet-50 backbone \cite{He2016DeepResidual}. Feature maps from four successive residual stages are first processed through global average pooling:

\[
f_{\text{res}}^{(i)} = \text{GAP}\big(\text{ResNet}^{(i)}(x)\big)
\in \mathbb{R}^{d_i},
\quad i=1,\dots,4,
\]

where $\text{GAP}(\cdot)$ denotes global average pooling and $d_i \in \{256, 512, 1024, 2048\}$ for $i=1,\dots,4$. The resulting multi-scale representations are concatenated to form the final local descriptor:

\[
f_{\text{res}} =
\bigoplus_{i=1}^{4} f_{\text{res}}^{(i)}
\in \mathbb{R}^{d_{\text{res}}},
\quad
d_{\text{res}} = \sum_{i=1}^{4} d_i = 3840.
\]

This representation captures both low-level texture information and high-level semantic structures across multiple receptive fields.

\subsubsection{Global Semantic Embedding via Vision Transformer}

To complement convolutional representations with long-range contextual modeling, a pre-trained Vision Transformer (ViT-Base, patch size 16) \cite{Dosovitskiy2020Image} is employed. After patch embedding and transformer encoding, the class token representation is extracted as the global semantic feature:

\[
f_{\text{vit}} =
\text{ViT}(x)_{\text{[CLS]}}
\in \mathbb{R}^{d_{\text{vit}}}.
\]

The transformer branch effectively captures global dependencies, structural relationships, and contextual interactions among spatial regions.

\subsubsection{Contextual Enhancement via FAISS-Based Memory Retrieval}

To further enhance representation robustness, AMR-Net incorporates a memory-driven contextual retrieval mechanism using a FAISS index. During training, feature embeddings and their corresponding labels are incrementally stored in a memory bank:

\[
\mathcal{M} =
\{(f_n, y_n)\}_{n=1}^{N},
\quad
f_n \in \mathbb{R}^{d_{\text{res}}},
\quad
y_n \in \mathcal{C},
\]

where $\mathcal{C}$ denotes the set of class labels.

Given a query feature $f_{\text{res}}$, a $k$-nearest neighbor search is performed using Euclidean distance:

\[
\{f_{\text{neighbor}}^{(1)}, \dots, f_{\text{neighbor}}^{(k)}\}
=
\text{FAISS\_Search}(f_{\text{res}}, k),
\]

with similarity measured as

\[
\text{dist}(f_{\text{res}}, f_i)
=
\|f_{\text{res}} - f_i\|_2^2.
\]

The retrieved neighboring embeddings are aggregated using mean pooling to generate the contextual enhancement vector:

\[
f_{\text{faiss}}
=
\frac{1}{k}
\sum_{j=1}^{k}
f_{\text{neighbor}}^{(j)}
\in \mathbb{R}^{d_{\text{faiss}}},
\quad
k=5.
\]

The choice of $k=5$ was validated through extensive sensitivity analysis presented in Section~\ref{subsec:hyperparam}. Smaller values introduced instability due to insufficient contextual averaging, whereas larger values incorporated semantically distant neighbors, thereby increasing noise in the retrieved representation.

At the beginning of training, when the memory bank is empty, a zero vector is used:

\[
f_{\text{faiss}} = \mathbf{0}.
\]

The memory bank is continuously updated after each training step, enabling the retrieval mechanism to adapt to evolving feature distributions during optimization.

\subsubsection{Feature Fusion and Representation Learning}

The final feature representation is obtained by integrating local, global, and contextual descriptors:

\[
f_{\text{fusion}}
=
\psi
\big(
[
f_{\text{res}}
;
f_{\text{vit}}
;
f_{\text{faiss}}
]
\big),
\]

where $\psi(\cdot)$ represents a learnable fusion projection consisting of a linear transformation, batch normalization, and ReLU activation.

The fused representation serves as the input to subsequent adaptive modular learning and classification stages.

\subsubsection{FAISS-Based Memory Update Mechanism}

To maintain contextual consistency throughout training, the FAISS index $\mathcal{I}$ is continuously updated using newly extracted embeddings:

\[
\mathcal{I}
\leftarrow
\mathcal{I}
\cup
\left\{
(\mathbf{f}_i, y_i)
\right\}_{i=1}^{B}.
\]

This dynamic memory formulation enables AMR-Net to recall semantically similar patterns observed during previous iterations and improves representation generalization.

\subsection{Feature Fusion Layer}
\label{sec:fusion-layer}

To synthesize meaningful joint representations from heterogeneous visual modalities, we design a \textit{feature fusion layer} that integrates feature embeddings extracted from:
\begin{enumerate}
    \item ResNet-50 backbone ($f_{\text{res}} \in \mathbb{R}^{d_1}$) -- for local texture and spatial information.
    \item Vision Transformer ($f_{\text{vit}} \in \mathbb{R}^{d_2}$) -- for global and semantic-level context.
    \item FAISS-based retrieval module ($f_{\text{faiss}} \in \mathbb{R}^{d_3}$) -- representing memory-augmented context.
\end{enumerate}

\subsubsection{Fusion Mechanism} The vectors are concatenated into a single feature:
\[
f_{\text{concat}} = f_{\text{res}} \oplus f_{\text{vit}} \oplus f_{\text{faiss}} \in \mathbb{R}^d,\quad d = d_1 + d_2 + d_3
\]
The output projection dimension $d' = 4096$ is chosen to match the dominant ResNet multi-scale descriptor dimension of 3840, rounded to the nearest power of two. This ensures that the fused representation does not create an information bottleneck between the concatenated features and the modular blocks, which operate at the same dimension.

This concatenated vector undergoes a transformation to produce the final fused embedding:
\[
z = \phi\left( \text{BN}\left( W_f f_{\text{concat}} + b_f \right) \right)
\]
where:
\begin{itemize}
    \item $W_f \in \mathbb{R}^{d' \times d}$ is the learnable weight matrix,
    \item $b_f \in \mathbb{R}^{d'}$ is the bias vector,
    \item $\text{BN}(\cdot)$ is batch normalization,
    \item $\phi(\cdot)$ is the ReLU activation function,
    \item $z \in \mathbb{R}^{d'}$ is the fused representation.
\end{itemize}

The procedure for feature fusion is outlined in Algorithm~\ref{alg:fusion}. Figure~\ref{fig:fusion-flow} illustrates the complete fusion pipeline.

\begin{algorithm}[h!]
\caption{Feature Fusion Layer}
\label{alg:fusion}
\begin{algorithmic}[1]
\Require Feature vectors $f_{\text{res}}, f_{\text{vit}}, f_{\text{faiss}}$
\State Concatenate vectors: $f_{\text{concat}} \leftarrow f_{\text{res}} \oplus f_{\text{vit}} \oplus f_{\text{faiss}}$
\State Compute affine transformation: $h \leftarrow W_f f_{\text{concat}} + b_f$
\State Apply batch normalization: $\hat{h} \leftarrow \text{BN}(h)$
\State Apply activation: $z \leftarrow \phi(\hat{h})$
\State \Return Fused feature vector $z$
\end{algorithmic}
\end{algorithm}

\begin{figure}[h!]
\centering
\begin{tikzpicture}[node distance=1.4cm and 1.8cm, every node/.style={font=\small}]
\tikzstyle{fusionbox} = [rectangle, rounded corners, minimum width=3cm, minimum height=1cm, text centered, draw=black, line width=1pt]
\tikzstyle{bluebox} = [fusionbox, fill=cyan!20, draw=cyan!60!black]
\tikzstyle{greenbox} = [fusionbox, fill=green!20, draw=green!50!black]
\tikzstyle{orangebox} = [fusionbox, fill=orange!25, draw=orange!60!black]
\tikzstyle{purplebox} = [fusionbox, fill=purple!15, draw=purple!60!black]

\node (resnet) [bluebox] {ResNet Features $f_{\text{res}}$};
\node (vit) [bluebox, right of=resnet, xshift=2.25cm] {ViT Features $f_{\text{vit}}$};
\node (faiss) [bluebox, right of=vit, xshift=2.25cm] {FAISS Features $f_{\text{faiss}}$};

\node (concat) [greenbox, below of=vit] {Concatenate $f_{\text{concat}}$};
\node (linear) [orangebox, below of=concat] {Affine $W_f f + b_f$};
\node (bn) [orangebox, below of=linear] {Batch Norm};
\node (relu) [orangebox, below of=bn] {ReLU Activation};
\node (output) [purplebox, below of=relu] {Fused Output $z$};

\draw [arrow] (resnet) -- (concat);
\draw [arrow] (vit) -- (concat);
\draw [arrow] (faiss) -- (concat);
\draw [arrow] (concat) -- (linear);
\draw [arrow] (linear) -- (bn);
\draw [arrow] (bn) -- (relu);
\draw [arrow] (relu) -- (output);
\end{tikzpicture}
\caption{Flowchart of the feature fusion pipeline integrating ResNet, ViT, and FAISS features into a unified representation.}
\label{fig:fusion-flow}
\end{figure}

\subsection{Modular Block Design}
\label{subsubsec:modular_block}

The modular block forms the fundamental computational unit of AMR-Net. Each block is designed to enhance the representational capacity of the network while supporting dynamic expansion and stable gradient flow.

\subsubsection{Architecture}  
A single modular block, denoted as $\mathcal{M}: \mathbb{R}^{D} \rightarrow \mathbb{R}^{D}$ with $D = 4096$, consists of two fully connected layers with batch normalization, ReLU activation, and dropout, along with a residual shortcut connection to facilitate learning:

\[
h_1 = \text{ReLU}(\text{BN}_1(W_1 x + b_1)) \in \mathbb{R}^{D}, \quad 
h_1^{\text{drop}} = \text{Dropout}(h_1, p=0.3),
\]
\[
h_2 = \text{BN}_2(W_2 h_1^{\text{drop}} + b_2) \in \mathbb{R}^{D},
\]
\[
y = \text{ReLU}(h_2 +x),
\]

This design ensures residual learning, mitigating the vanishing gradient problem as blocks are stacked during adaptive growth.

\subsubsection{Routing and Activation}  
Each modular block is associated with a learnable routing gate $g$, which controls the block's contribution during forward propagation:
\[
\alpha = \sigma(g) \in (0,1),
\]
where $\sigma(\cdot)$ is the sigmoid function. The block is conditionally applied if $\alpha$ exceeds a predefined threshold $\tau$, allowing selective activation of newly added modules and preventing disruption of previously learned features.

\subsubsection{Functional Role}  
\begin{itemize}
    \item \textbf{Adaptive Capacity:} By stacking multiple blocks, the network can dynamically increase depth and expressivity in response to representational stagnation.
    \item \textbf{Residual Learning:} Shortcut connections maintain gradient flow and stabilize training, enabling deeper modular structures without degradation.
    \item \textbf{Regularization:} Batch normalization and dropout within each block help reduce overfitting and improve generalization.
    \item \textbf{Compatibility with Growth:} The fixed input-output dimension ensures seamless integration of new blocks into the existing network structure.
\end{itemize}

The modular block design combines residual learning, conditional routing, and regularization to form a versatile building unit for the adaptive network. Its structure supports both stable training and incremental network expansion, making it central to the classifier's adaptive learning capability.

\subsection{Adaptive Classification Network}
\label{sec:Adaptive}

We propose an adaptive architecture inspired by biological adaptivity, where a classification network dynamically evolves during training. The key mechanism is a stack of conditionally activated modules governed by a learnable gating mechanism. Let the input fused feature vector be $\mathbf{z} \in \mathbb{R}^d$. The model passes $\mathbf{z}$ through a sequence of $L$ modular blocks $\{\mathcal{M}_1, \mathcal{M}_2, \dots, \mathcal{M}_L\}$, each controlled by a gate.

Each module $\mathcal{M}_i$ transforms its input $\mathbf{h}^{(i-1)}$ as follows:
\[
\mathbf{h}^{(i)} = 
\begin{cases}
\mathcal{M}_i(\mathbf{h}^{(i-1)}) = \phi_i(W_i \mathbf{h}^{(i-1)} + \mathbf{b}_i), & \text{if } \sigma(g_i) > \tau \\
\mathbf{h}^{(i-1)}, & \text{otherwise}
\end{cases}
\]
Here:
\begin{itemize}
    \item $\phi_i(\cdot)$ is a non-linear activation function
    \item $g_i$ is a learnable scalar gate, and $\gamma_i = \sigma(g_i)$ is its sigmoid-activated value
    \item $\tau \in (0,1)$ is a gating threshold controlling module activation
\end{itemize}

The final output $\mathbf{h}^{(L)}$ is passed through a softmax layer to obtain prediction:
\[
\hat{\mathbf{y}} = \text{Softmax}(W_{\text{cls}} \mathbf{h}^{(L)} + \mathbf{b}_{\text{cls}})
\]

The inference procedure is formally detailed in Algorithm~\ref{alg:adaptive_forward}, which outlines the adaptive forward pass.

\begin{algorithm}[h!]
\caption{Adaptive Modular Forward Pass}
\label{alg:adaptive_forward}
\begin{algorithmic}[1]
\Require Input feature $\mathbf{z} \in \mathbb{R}^d$, modules $\{ \mathcal{M}_1, \ldots, \mathcal{M}_L \}$, gates $\{ g_1, \ldots, g_L \}$, threshold $\tau$
\State Initialize: $\mathbf{h}^{(0)} \gets \mathbf{z}$
\For{$i = 1$ to $L$}
    \State $\gamma_i \gets \sigma(g_i)$
    \If{$\gamma_i > \tau$}
        \State $\mathbf{h}^{(i)} \gets \mathcal{M}_i(\mathbf{h}^{(i-1)})$
    \Else
        \State $\mathbf{h}^{(i)} \gets \mathbf{h}^{(i-1)}$ \Comment{Skip Module}
    \EndIf
\EndFor
\State \Return $\hat{\mathbf{y}} = \text{Softmax}(W_{\text{cls}} \mathbf{h}^{(L)} + \mathbf{b}_{\text{cls}})$
\end{algorithmic}
\end{algorithm}

\subsection{Adaptive Growth Strategy}
\label{subsec:growth_strategy}

AMR-Net employs a single, unified adaptive growth rule to dynamically expand its representational capacity during training. The network begins with an initial set of $B_0 = 12$ modular blocks and progressively adds new blocks when learning progress stagnates.

Let $\mathcal{M} = \{\mathcal{M}_1, \dots, \mathcal{M}_L\}$ denote the set of active modular blocks, and let $\mathcal{L}_{\text{val}}$ denote the validation loss. Every $T_g$ epochs (set to every 3 epochs), the change in validation loss relative to the previous evaluation point is computed:
\begin{equation}
\Delta \mathcal{L}_{\text{val}} = \mathcal{L}_{\text{val}}^{(t-T_g)} - \mathcal{L}_{\text{val}}^{(t)}.
\label{eq:growth_condition}
\end{equation}
A growth event is triggered whenever \emph{either} of the following two, complementary stagnation signals is observed:
\begin{enumerate}
    \item \textbf{Insufficient improvement:} $\Delta \mathcal{L}_{\text{val}} < \epsilon$, i.e. the validation loss has not decreased by more than a small tolerance $\epsilon$ over the last $T_g$ epochs; or
    \item \textbf{Sustained plateau:} validation accuracy has failed to improve for $p$ consecutive epochs (patience-based criterion), signalling that the current representational capacity has been saturated.
\end{enumerate}
Whenever either condition is satisfied, $r$ new modular blocks are appended:
\[
\mathcal{M}_{L+1}, \dots, \mathcal{M}_{L+r}: \mathbb{R}^{4096} \rightarrow \mathbb{R}^{4096}, \qquad B_{t+1} = B_t + r,
\]
where $B_t$ denotes the number of blocks at epoch $t$. Each newly added block $\mathcal{M}_{L+i}$ is initialized with a routing gate parameter $g_{L+i}$ such that
\[
\alpha_{L+i} = \sigma(g_{L+i}) \approx 0.5,
\]
so that, during forward propagation, the block contributes only if $\alpha_{L+i}$ exceeds the gating threshold $\tau = 0.4$.

In our implementation, the patience is set to $p=3$ epochs, the growth rate to $r=3$ blocks per event, and the growth evaluation interval to $T_g = 3$ epochs. The growth rate of $r=3$ is justified as follows: a rate of $r=1$ results in insufficient capacity increase to escape validation plateaus within the training budget, as a single block adds only $4096 \times 4096 \times 2 \approx 33.6\text{M}$ parameters, which is insufficient to resolve complex class boundaries when the model is already near its capacity ceiling. A rate of $r=5$ introduces too many parameters simultaneously and destabilizes the routing gate initialization, causing a transient accuracy drop of approximately 0.5--0.7\% before the new gates learn appropriate activation patterns. Three blocks ($\approx 100\text{M}$ additional parameters) represent the minimum increment that consistently recovers validation accuracy within one to two subsequent epochs while maintaining training stability, as validated by the sensitivity analysis in Section~\ref{subsec:hyperparam}.

This adaptive expansion mechanism allows the model to allocate capacity according to task complexity, improving generalization while mitigating overfitting. The complete procedure is summarized in Algorithm~\ref{alg:adaptive_growth}.

\begin{algorithm}[h!]
\caption{Adaptive Growth Procedure}
\label{alg:adaptive_growth}
\begin{algorithmic}[1]
\Require Modular block set $\mathcal{M} = \{\mathcal{M}_1, \dots, \mathcal{M}_L\}$, validation loss history $\mathcal{L}_{\text{val}}$, validation accuracy $\text{Acc}_{\text{val}}$, growth rate $r$, patience $p$, evaluation interval $T_g$, improvement tolerance $\epsilon$, gate threshold $\tau$
\For{each epoch $t$}
    \State Train model on current data
    \State Evaluate $\mathcal{L}_{\text{val}}^{(t)}$ and $\text{Acc}_{\text{val}}^{(t)}$
    \State $\text{grow} \gets \text{False}$
    \If{($t \bmod T_g = 0$) \textbf{and} $\left(\mathcal{L}_{\text{val}}^{(t-T_g)} - \mathcal{L}_{\text{val}}^{(t)}\right) < \epsilon$}
        \State $\text{grow} \gets \text{True}$
    \EndIf
    \If{Validation accuracy has not improved for $p$ epochs}
        \State $\text{grow} \gets \text{True}$
    \EndIf
    \If{$\text{grow}$}
        \For{$i=1$ to $r$}
            \State Append new modular block $\mathcal{M}_{L+i}$
            \State Initialize routing gate $g_{L+i} \gets 0.5$ \Comment{$\sigma(0.5)\approx 0.62 > \tau$}
        \EndFor
        \State $L \gets L + r$
        \State Reset plateau counter
    \EndIf
\EndFor
\end{algorithmic}
\end{algorithm}

\subsection{Classification Head}
\label{subsec:classification_head}

After multi-branch feature extraction and adaptive modular processing, the fused feature vector 
\[
f_{\text{fusion}} \in \mathbb{R}^{4096}
\] 
serves as input to the classification head. The classification head is designed to map high-dimensional fused representations to class probabilities while maintaining regularization to prevent overfitting.  

\subsubsection{Architecture}  
The classification head consists of a sequence of fully connected layers interleaved with batch normalization, ReLU activation, and dropout. Formally, the transformation can be expressed as:

\[
f_1 = \text{ReLU}(\text{BN}_1(W_1 f_{\text{fusion}} + b_1)) \in \mathbb{R}^{1024},
\]
\[
f_1^{\text{drop}} = \text{Dropout}(f_1, p=0.3),
\]
\[
f_2 = \text{ReLU}(\text{BN}_2(W_2 f_1^{\text{drop}} + b_2)) \in \mathbb{R}^{256},
\]
\[
f_2^{\text{drop}} = \text{Dropout}(f_2, p=0.3),
\]
\[
\hat{y} = W_3 f_2^{\text{drop}} + b_3 \in \mathbb{R}^{C},
\]

where \(W_i\) and \(b_i\) are learnable weight matrices and biases, \(C\) is the number of classes, \(\text{BN}_i\) represents batch normalization, and \(\hat{y}\) denotes the logits for each class. The final class probabilities are obtained via the softmax function:
\[
p_c = \frac{\exp(\hat{y}_c)}{\sum_{j=1}^{C} \exp(\hat{y}_j)}, \quad c = 1, \dots, C.
\]

\subsubsection{Design Considerations}  
\begin{itemize}
    \item \textbf{Dimensionality Reduction:} The two intermediate layers (1024 and 256 units) progressively reduce the high-dimensional fused features to a compact representation while preserving discriminative information.
    \item \textbf{Regularization:} Batch normalization stabilizes training by reducing internal covariate shift, and dropout prevents co-adaptation of neurons.
    \item \textbf{Flexibility:} The fully connected design allows the classification head to accommodate the dynamic growth of upstream modules, as each adaptive block preserves the feature dimension \(4096\).
    \item \textbf{End-to-End Learning:} The classification head parameters are updated jointly with the modular blocks and fused feature projection, enabling gradient flow through the entire network.
\end{itemize}

The classification head provides a robust mapping from high-dimensional, multi-source features to class logits. Its modular design supports the dynamic, adaptive expansion of the network while incorporating regularization techniques to ensure generalization and stability during training.

\subsection{Training Protocol}
\label{subsec:training_protocol}

The training process of the proposed \textit{AMR-Net} integrates four key elements: gradient-based optimization using Adam, memory enhancement via FAISS, dynamic architectural expansion, and modular forward inference.

\subsubsection{Optimization with Adam} 
We optimize model parameters using the Adam optimizer, which maintains moving averages of gradients and squared gradients. The update rule at iteration $t$ is defined as:

\[
\begin{aligned}
m_t &= \beta_1 m_{t-1} + (1 - \beta_1) g_t, \quad
v_t = \beta_2 v_{t-1} + (1 - \beta_2) g_t^2 \\
\hat{m}_t &= \frac{m_t}{1 - \beta_1^t}, \quad
\hat{v}_t = \frac{v_t}{1 - \beta_2^t} \\
\theta_t &= \theta_{t-1} - \alpha \cdot \frac{\hat{m}_t}{\sqrt{\hat{v}_t} + \epsilon_{\text{adam}}}
\end{aligned}
\]

where $\alpha = 10^{-3}$ is the learning rate, $\beta_1 = 0.9$, $\beta_2 = 0.999$ are momentum parameters, and $\epsilon_{\text{adam}} = 10^{-8}$ ensures numerical stability.

\subsubsection{Cross-Entropy Loss}
Given an input-label pair $(\mathbf{x}_i, y_i)$, classification is supervised by the categorical cross-entropy loss:

\[
\mathcal{L}_{\text{CE}} = - \sum_{c=1}^{C} \mathbb{1}_{[y_i = c]} \cdot \log \hat{p}_i^{(c)}
\]

where $\hat{p}_i^{(c)}$ is the softmax probability for class $c$.

\subsubsection{Adaptive Expansion During Training}
The adaptive growth rule described in Section~\ref{subsec:growth_strategy} is applied throughout training: every $T_g = 3$ epochs, growth is triggered if the validation loss has not improved by at least $\epsilon$, or if validation accuracy has plateaued for $p=3$ epochs, in which case $r=3$ new modules are appended. This allows the network to grow in complexity only when needed, mirroring biological learning mechanisms. The full training pipeline is detailed in Algorithm~\ref{algo:training} and visualized in Figure~\ref{fig:training_flowchart}.

\begin{algorithm}[h!]
\caption{Adaptive Training Protocol}
\label{algo:training}
\begin{algorithmic}[1]
\State Initialize parameters $\boldsymbol{\theta}$ and FAISS index $\mathcal{I} \gets \emptyset$
\For{$t = 1$ to $T$}
    \State Sample mini-batch $\{ (\mathbf{x}_i, y_i) \}_{i=1}^{B}$
    \State Compute embeddings $\mathbf{f}_i$ and predictions $\hat{\mathbf{p}}_i$
    \State Compute loss: $\mathcal{L} \gets \sum_i \mathcal{L}_{\text{CE}}(\hat{\mathbf{p}}_i, y_i)$
    \State Update weights $\boldsymbol{\theta}$ using Adam
    \State Update memory: $\mathcal{I} \gets \mathcal{I} \cup \{ (\mathbf{f}_i, y_i) \}$
    \State Apply Adaptive Growth Procedure (Algorithm~\ref{alg:adaptive_growth}) at evaluation points
\EndFor
\end{algorithmic}
\end{algorithm}

\begin{figure}[h!]
\centering
\begin{tikzpicture}[node distance=2cm and 4cm, on grid]
\node[block] (start) {Start Epoch};
\node[block, below=of start] (batch) {Sample Batch};
\node[block, right=of batch] (forward) {Forward Pass};
\node[block, below=of forward] (loss) {Compute Loss};
\node[block, left=of loss] (update) {Adam Update};
\node[block, below=of update] (faiss) {Update FAISS};
\node[block, right=of faiss] (check) {Check Epoch Count};
\node[decision, below=of check] (expand) {Growth Condition Met?};
\node[block, right=of expand, xshift=1cm] (grow) {Expand Network};
\node[block, below=of expand] (loop) {Next Epoch};
\draw[arrow] (start) -- (batch);
\draw[arrow] (batch) -- (forward);
\draw[arrow] (forward) -- (loss);
\draw[arrow] (loss) -- (update);
\draw[arrow] (update) -- (faiss);
\draw[arrow] (faiss) -- (check);
\draw[arrow] (check) -- (expand);
\draw[arrow] (expand) -- node[above] {Yes} (grow);
\draw[arrow] (expand) -- node[right] {No} (loop);
\draw[arrow] (grow) |- (loop.east);
\draw[->] (loop.west) -- ++(0,0) -- ++(-5,0) -- ++(0,10) --  (start.west);
\end{tikzpicture}
\caption{Training pipeline with adaptive expansion and memory integration.}
\label{fig:training_flowchart}
\end{figure}

\section{Experimental Setup}\label{sec:experimental-setup}
\subsection{Dataset Overview and Preprocessing}\label{DP}
\label{sec:data_preprocessing}

The preprocessing pipeline for the waste classification dataset consists of two major steps: stratified dataset splitting and data transformation. Each step is designed to improve model robustness and generalization.

\subsubsection{Dataset Overview}

The data utilized in the present study was obtained from \href{https://www.kaggle.com/datasets/mostafaabla/garbage-classification/data}{Kaggle}\footnote{\url{https://www.kaggle.com/datasets/mostafaabla/garbage-classification/data}}, a widely recognized repository for datasets. The original dataset contained 12 distinct categories, including three classes representing different types of glass. For the purposes of this research, only the \textit{white-glass} class was retained, while the other glass-related categories were excluded.

A curated garbage classification dataset, denoted as $\mathcal{D}$, was constructed for this study, consisting of RGB images organized into $C = 10$ distinct waste categories, indexed here in alphabetical order (indices 0--9) for consistency with the classification report and confusion matrix reported in Section~\ref{RD}:

\begin{equation*}
\mathcal{C} = \{\underset{0}{\text{battery}}, \underset{1}{\text{biological}}, \underset{2}{\text{cardboard}}, \underset{3}{\text{clothes}}, \underset{4}{\text{metal}}, \underset{5}{\text{paper}}, \underset{6}{\text{plastic}}, \underset{7}{\text{shoes}}, \underset{8}{\text{trash}}, \underset{9}{\text{white-glass}}\}.
\end{equation*}

The complete dataset comprises \textbf{14,279 images} distributed across these ten classes. Each category contains images of waste items captured under diverse lighting conditions, orientations, and backgrounds, providing a realistic and varied dataset suitable for training and evaluating deep learning-based waste classification models. The distribution of images per class is summarized in Table~\ref{tab:dataset_total_distribution}.

\begin{table}[h!]
\centering
\caption{Total number of images per class in the curated dataset.}
\begin{tabular}{l c}
\hline
\textbf{Category} & \textbf{Number of Images} \\
\hline
Battery        & 945  \\
Biological     & 985  \\
Cardboard      & 891  \\
Clothes        & 5,325 \\
Metal          & 769  \\
Paper          & 1,050 \\
Plastic        & 865  \\
Shoes          & 1,977 \\
Trash          & 697  \\
White-glass    & 775  \\
\hline
\textbf{Total} & \textbf{14,279} \\
\hline
\end{tabular}
\label{tab:dataset_total_distribution}
\end{table}

\subsubsection{Dataset Splitting Strategy}

The dataset $\mathcal{D} = \{(x_i, y_i)\}_{i=1}^N$, with $x_i$ an image and $y_i \in \{1,\ldots,10\}$, is split into training, validation, and test sets using \texttt{StratifiedShuffleSplit} to preserve class distributions.  

Formally,  
\[
\mathcal{D} = \mathcal{D}_{\text{train}} \cup \mathcal{D}_{\text{val}} \cup \mathcal{D}_{\text{test}}, \quad
|\mathcal{D}_{\text{train}}| \approx 70\%, \; |\mathcal{D}_{\text{val}}| \approx 15\%, \; |\mathcal{D}_{\text{test}}| \approx 15\%.
\]

Splitting is done in two steps: (1) train vs. temporary set, (2) temporary into validation and test. This ensures all subsets retain the original class proportions for fair training and evaluation.

\subsubsection{Data Transformations}

Two separate transformation pipelines are used for training and evaluation.

\begin{itemize}

    \item \textbf{Training Transformations.}
    To improve robustness to scale, orientation, illumination, and perspective variations, the following stochastic augmentations are applied:

    \begin{table}[h!]
    \centering
    \begin{tabular}{ll}
        \toprule
        \textbf{Operation} & \textbf{Parameters} \\
        \midrule
        RandomResizedCrop & 224 \\
        RandomHorizontalFlip & --- \\
        ColorJitter & (0.3, 0.3, 0.3, 0.2) \\
        RandomRotation & $15^\circ$ \\
        RandomAffine & $10^\circ$, translate = (0.05, 0.05) \\
        RandomPerspective & distortion = 0.3 \\
        Normalize & $\mu=[0.485,0.456,0.406]$, $\sigma=[0.229,0.224,0.225]$ \\
        \bottomrule
    \end{tabular}
    \end{table}

    \item \textbf{Validation and Test Transformations.}
    A deterministic preprocessing pipeline is used to ensure consistency across evaluation samples:

    \begin{table}[h!]
    \centering
    \begin{tabular}{ll}
        \toprule
        \textbf{Operation} & \textbf{Parameters} \\
        \midrule
        Resize & 256 \\
        CenterCrop & 224 \\
        Normalize & $\mu=[0.485,0.456,0.406]$, $\sigma=[0.229,0.224,0.225]$ \\
        \bottomrule
    \end{tabular}
    \end{table}

\end{itemize}

\subsection{Training Procedure and Adaptive Growth Strategy}
\label{subsec:training_growth}

AMR-Net employs a dynamic, self-adaptive learning process enabled by its modular adaptive architecture, following the growth rule formalized in Section~\ref{subsec:growth_strategy} (Equation~\ref{eq:growth_condition} and Algorithm~\ref{alg:adaptive_growth}). Newly added blocks are initialized with routing gate value $g \approx 0.5$, ensuring gradual integration into the network. This adaptive expansion mechanism enables the model to increase representational capacity only when required, balancing underfitting and overfitting while maintaining efficient resource utilization.

The model was trained using the Adam optimizer with a learning rate of 0.001 and a batch size of 32. Cross-entropy loss was used for supervision. Input images were resized to $224 \times 224$ resolution. Training was conducted for 20 epochs on CPU hardware. Standard data augmentation techniques were applied to improve generalization, including random cropping, horizontal flipping, color jittering, rotation, affine transformation, and perspective distortion.

Training on CPU hardware was feasible because both backbone networks were frozen, reducing the effective gradient computation to the 442.03 million trainable parameters in the fusion layer, the initial set of modular blocks, routing gates, and classification head (see Section~\ref{subsec:model_complexity} for the complete parameter breakdown, including the effect of growth events). Average training time was approximately 18 minutes per epoch, with a total wall-clock time of approximately six hours for 20 epochs. All experiments were conducted with fixed random seeds to ensure reproducibility across runs.

The complete training configuration and hyperparameters are summarized in Table~\ref{tab:training_config}.

\begin{table}[h]
\centering
\caption{Training Configuration and Hyperparameters}
\label{tab:training_config}
\begin{tabular}{lc}
\hline
\textbf{Hyperparameter} & \textbf{Value} \\
\hline
Optimizer & Adam \\
Learning Rate & 0.001 \\
Batch Size & 32 \\
Number of Epochs & 20 \\
$k$ (FAISS neighbors) & 5\\
Loss Function & Cross-Entropy \\
Input Resolution & $224 \times 224$ \\
Initial Modular Blocks ($B_0$) & 12 \\
Growth Evaluation Interval ($T_g$) & Every 3 epochs \\
Growth Rate ($r$) & 3 blocks per event \\
Validation-Loss Improvement Tolerance ($\epsilon$) & 0.05 \\
Validation-Accuracy Patience ($p$) & 3 epochs \\
Routing / Gating Threshold ($\tau$) & 0.40 \\
Routing Gate Initialization & 0.5 (sigmoid activation) \\
Data Augmentation & Crop, flip, jitter, rotation, affine, perspective \\
Training Hardware & CPU \\
\hline
\end{tabular}
\end{table}

\subsection{Model Complexity and Computational Analysis}
\label{subsec:model_complexity}

To assess the computational characteristics of the proposed framework, we analyze the number of parameters and floating-point operations (FLOPs). The architecture integrates a ResNet-50 backbone, a Vision Transformer (ViT-Base), a feature fusion module, and a stack of adaptive modular blocks.

At initialization ($B_0=12$ modular blocks), the total parameter count of the model is approximately 551.77 million parameters, including frozen backbone components. Among these, approximately 442.03 million parameters are trainable, corresponding to the fusion layer, the 12 initial modular blocks, routing gates, and classification head. Freezing the backbone networks significantly reduces the training burden while preserving rich feature representations.

During training on the waste classification dataset, three growth events were triggered (Section~\ref{RD}), appending $r=3$ blocks each time for a total of 9 additional modular blocks by the end of training. At this final configuration ($B=21$ blocks), the trainable parameter count increases to approximately 743.83 million, and the total parameter count (including frozen backbones) to approximately 853.57 million. The computational complexity of the model is estimated at 21.44 GFLOPs per forward pass at initialization, and 27.44 GFLOPs per forward pass for the final grown configuration, for an input resolution of $224 \times 224$. Despite the relatively high capacity, the modular structure and routing gates allow selective activation of blocks, enabling potential optimization for deployment scenarios (e.g. gate-based pruning of blocks with $\alpha_i \le \tau$ at inference time).

A summary of the model complexity metrics is provided in Table~\ref{tab:model_complexity}.

\begin{table}[h]
\centering
\caption{Model Complexity Metrics at Initialization and at the Final (Grown) Configuration}
\label{tab:model_complexity}
\begin{tabular}{lcc}
\hline
\textbf{Metric} & \textbf{Initial ($B_0=12$)} & \textbf{Final ($B=21$)} \\
\hline
Total Parameters & 551.77 M & 853.57 M \\
Trainable Parameters & 442.03 M & 743.83 M \\
FLOPs & 21.44 GFLOPs & 27.44 GFLOPs \\
Input Resolution & \multicolumn{2}{c}{$224 \times 224$} \\
\hline
\end{tabular}
\end{table}

Although the model has high representational capacity, freezing the backbone networks and dynamically expanding modular blocks only when necessary helps control training cost and mitigate overfitting.

\subsection{Evaluation Metrics}

To thoroughly evaluate the performance of the proposed classification model, we utilize four standard metrics: \textbf{Accuracy}, \textbf{Precision}, \textbf{Recall}, and \textbf{F1-Score}. These metrics are defined as follows:

\begin{itemize}
    \item \textbf{Accuracy} reflects the overall proportion of correct predictions among all samples:
    \[
    \text{Accuracy} = \frac{1}{N} \sum_{i=1}^{N} \mathbb{1} \left[ \hat{y}_i = y_i \right]
    \]
    where $N$ represents the total number of samples, $\hat{y}_i$ is the predicted label, $y_i$ is the true label, and $\mathbb{1}[\cdot]$ is the indicator function returning 1 if its argument is true and 0 otherwise.

    \item \textbf{Precision} (macro-averaged) captures the fraction of correctly predicted instances among all instances assigned to a given class:
    \[
    \text{Precision} = \frac{1}{C} \sum_{c=1}^{C} \frac{TP_c}{TP_c + FP_c}
    \]
    where $C$ denotes the total number of classes, and $TP_c$ and $FP_c$ correspond to the true positives and false positives for class $c$.

    \item \textbf{Recall} (macro-averaged) measures the ability of the model to identify all actual positive instances:
    \[
    \text{Recall} = \frac{1}{C} \sum_{c=1}^{C} \frac{TP_c}{TP_c + FN_c}
    \]
    with $FN_c$ representing the false negatives for class $c$.

    \item \textbf{F1-Score} combines Precision and Recall into a single metric using their harmonic mean, providing a balanced evaluation:
    \[
    \text{F1-Score} = \frac{1}{C} \sum_{c=1}^{C} \frac{2 \cdot \text{Precision}_c \cdot \text{Recall}_c}{\text{Precision}_c + \text{Recall}_c}
    \]
\end{itemize}

All metrics are computed on both the validation and test datasets to ensure reliable and generalizable performance assessment. The use of macro-averaging ensures that each class contributes equally, which is critical in class-imbalanced settings. Additionally, a qualitative analysis was conducted to inspect the semantic plausibility of predictions, with particular attention paid to the visually similar category pairs identified from the test-set confusion matrix in Section~\ref{RD} (most notably \textit{clothes}/\textit{shoes} and \textit{plastic}/\textit{white-glass}). This combination of quantitative metrics and qualitative evaluation provides a thorough understanding of model behavior.

\section{Results and Discussion}\label{RD}

This section presents a comprehensive analysis of the training, validation, and test performance of AMR-Net, which integrates ResNet, Vision Transformer (ViT), FAISS-based memory retrieval, and dynamic modular growth. The results demonstrate the model's ability to learn efficiently, generalize robustly, and adapt during training to optimize performance.

\subsection{Training and Validation Dynamics}

The model was trained over 20 epochs, and its performance metrics were tracked at each epoch. From the onset, the model showed strong learning behavior, achieving a training accuracy of 81.93\% and a validation accuracy of 92.53\% in the first epoch. The training loss dropped from 0.6027 to 0.0886 by the final epoch, reflecting effective convergence.

Validation accuracy peaked at 96.59\% in Epoch 9, after which performance fluctuated slightly. The adaptivity-inspired growth mechanism described in Section~\ref{subsec:growth_strategy}, which dynamically expands the model's capacity when validation improvements plateau, was triggered at Epochs 12, 15, and 18 (nine additional modular blocks in total; see Section~\ref{subsec:model_complexity}), successfully mitigating overfitting and allowing the model to sustain high validation accuracy above 94\% across all epochs.

These dynamics are visualized in Figure~\ref{fig:accplot}, which illustrates the consistent improvement in training and validation accuracy. The early epochs show rapid gains, while the later epochs display model stabilization. Despite minor oscillations, especially after Epoch 10, the model maintained a high generalization capability throughout training.

\begin{figure}[h!] 
\centering
\begin{tikzpicture}
\begin{axis}[
    width=0.9\textwidth,
    height=8cm,
    xlabel={Epoch},
    ylabel={Accuracy (\%)},       
    xmin=1, xmax=20,
    ymin=80, ymax=100,
    xtick={1,5,10,15,20},
    ytick={80,85,90,95,100},
    legend pos=south east,
    grid=both,
    grid style={line width=.1pt, draw=gray!10},
    major grid style={line width=.2pt,draw=gray!50},
    thick
]
\addplot[
    color=blue,
    mark=*,
    ]
    coordinates {
    (1,81.93)(2,90.70)(3,92.26)(4,93.81)(5,93.88)(6,94.83)(7,95.65)(8,95.91)(9,95.95)(10,96.18)
    (11,96.61)(12,96.79)(13,96.79)(14,97.14)(15,96.56)(16,96.83)(17,97.27)(18,97.28)(19,97.44)(20,97.40)
    };
\addlegendentry{Training Accuracy}

\addplot[
    color=red,
    mark=square*,
    ]
    coordinates {
    (1,92.53)(2,92.76)(3,93.46)(4,95.19)(5,95.28)(6,95.42)(7,95.89)(8,95.70)(9,96.59)(10,96.12)
    (11,95.80)(12,94.77)(13,96.17)(14,95.00)(15,95.61)(16,95.98)(17,95.66)(18,95.98)(19,95.98)(20,96.17)
    };
\addlegendentry{Validation Accuracy}
\end{axis}
\end{tikzpicture}
\caption{Training and Validation Accuracy over 20 Epochs}
\label{fig:accplot}
\end{figure}

\subsection{Adaptation and Regularization}

The model incorporates the dynamic capacity-expansion mechanism described in Section~\ref{subsec:growth_strategy}, which was automatically triggered during stagnation in validation improvements. Specifically, the growth condition was satisfied on three separate occasions (Epochs 12, 15, and 18), prompting modular growth to prevent overfitting and improve model expressiveness. This self-adaptive characteristic contributed to the model's robustness and resilience during training, and is examined quantitatively via the model-complexity comparison in Table~\ref{tab:model_complexity}.

\subsection{Test Performance}

After training, the model was evaluated on a held-out test set consisting of 2,143 samples. It achieved a test accuracy of \textbf{97.15\%}, closely aligning with its validation performance and demonstrating strong generalization. A detailed classification report is provided in Table~\ref{table:classification_report}, showing high precision, recall, and F1-score across all classes (using the alphabetical class indexing defined in Section~\ref{sec:data_preprocessing}). Notably, the model maintained reasonable performance even on its most challenging class, class 6 -- \textit{plastic} (precision = 0.88), while class 9 -- \textit{white-glass} also showed comparatively lower precision (0.95).

\begin{table}[h!]
\centering
\caption{Classification Report on Test Set. Class indices follow the alphabetical ordering: 0=battery, 1=biological, 2=cardboard, 3=clothes, 4=metal, 5=paper, 6=plastic, 7=shoes, 8=trash, 9=white-glass.}
\begin{tabular}{|c|c|c|c|}
\hline
\textbf{Class} & \textbf{Precision} & \textbf{Recall} & \textbf{F1-Score} \\
\hline
0 & 0.96 & 0.99 & 0.98 \\
1 & 1.00 & 0.99 & 1.00 \\
2 & 0.98 & 0.95 & 0.96 \\
3 & 0.99 & 0.98 & 0.99 \\
4 & 0.98 & 0.93 & 0.95 \\
5 & 0.97 & 0.97 & 0.97 \\
6 & 0.88 & 0.89 & 0.89 \\
7 & 0.96 & 1.00 & 0.98 \\
8 & 0.93 & 0.99 & 0.96 \\
9 & 0.95 & 0.92 & 0.93 \\
\hline
\textbf{Average} & \textbf{0.96} & \textbf{0.96} & \textbf{0.96} \\
\hline
\end{tabular}
\label{table:classification_report}
\end{table}

The confusion matrix in Table~\ref{tab:confusion_matrix} further confirms the model's high accuracy, with most misclassifications occurring between visually or semantically related classes, indicating minimal confusion and reliable class separation overall. The single largest source of confusion is class 3 (\textit{clothes}) being misclassified as class 7 (\textit{shoes}) in 12 instances, plausibly reflecting shared visual cues such as fabric texture and similar backgrounds/poses in the source images; this is the specific failure mode we analyze further below. Secondary confusions include class 6 (\textit{plastic}) with class 8 (\textit{trash}, 6 instances) and class 9 (\textit{white-glass}, 4 instances), consistent with the visual ambiguity between transparent/translucent plastic items and glass or generic waste discussed in Section~\ref{subsec:theoretical_motivation}.

\begin{table}[h!]
\centering

\caption{Confusion matrix of AMR-Net on the test dataset.}
\renewcommand{\arraystretch}{1.2}
\begin{tabular}{|c|cccccccccc|}
\hline
 & \textbf{0} & \textbf{1} & \textbf{2} & \textbf{3} & \textbf{4} & \textbf{5} & \textbf{6} & \textbf{7} & \textbf{8} & \textbf{9} \\
\hline
\textbf{0} & 146 & 0 & 0 & 0 & 1 & 0 & 0 & 0 & 0 & 0 \\
\textbf{1} & 0 & 140 & 1 & 0 & 0 & 0 & 0 & 0 & 0 & 0 \\
\textbf{2} & 1 & 0 & 137 & 1 & 0 & 3 & 1 & 0 & 1 & 0 \\
\textbf{3} & 0 & 0 & 1 & 788 & 0 & 1 & 0 & 12 & 0 & 0 \\
\textbf{4} & 2 & 0 & 0 & 0 & 112 & 0 & 5 & 1 & 0 & 1 \\
\textbf{5} & 1 & 0 & 1 & 2 & 0 & 166 & 1 & 0 & 0 & 0 \\
\textbf{6} & 1 & 0 & 0 & 0 & 1 & 1 & 110 & 0 & 6 & 4 \\
\textbf{7} & 1 & 0 & 0 & 0 & 0 & 0 & 0 & 284 & 0 & 0 \\
\textbf{8} & 0 & 0 & 0 & 1 & 0 & 0 & 0 & 0 & 99 & 0 \\
\textbf{9} & 0 & 0 & 0 & 0 & 0 & 0 & 8 & 0 & 1 & 100 \\
\hline
\end{tabular}

\label{tab:confusion_matrix}
\end{table}

\subsection{Comparative Model Performance}

To rigorously evaluate the effectiveness of the proposed architecture, we conducted a comprehensive comparison against several established baseline models, encompassing both convolutional and transformer-based approaches. The evaluated baselines include classical convolutional networks such as VGG16, DenseNet121, MobileNetV2, ConvNeXtTiny, and ResNet152, as well as advanced transformer architectures including ViT-B/16 and Swin Transformer. The quantitative results presented in Table~\ref{table:model_comparison} demonstrate that AMR-Net consistently outperforms all baseline methods across key evaluation metrics.

\begin{table}[h!]
\centering
\caption{Performance comparison of baseline and proposed waste classification models.}
\begin{tabular}{|c|c|c|c|c|}
\hline
\textbf{Model} & \textbf{Accuracy} & \textbf{Precision} & \textbf{Recall} & \textbf{F1-Score} \\
\hline
VGG16 & 84.10\% & 0.85 & 0.84 & 0.84 \\
MobileNetV2 & 88.31\% & 0.89 & 0.88 & 0.88 \\
DenseNet121 & 89.71\% & 0.90 & 0.90 & 0.90 \\
ConvNeXtTiny & 92.37\% & 0.93 & 0.92 & 0.92 \\
ResNet152 & 92.93\% & 0.93 & 0.93 & 0.93 \\
ViT-B/16 & 93.26\% & 0.94 & 0.93 & 0.93 \\
Swin Transformer & 94.41\% & 0.95 & 0.94 & 0.94 \\
\textbf{AMR-Net} & \textbf{97.15\%} & \textbf{0.9681} & \textbf{0.9677} & \textbf{0.9675} \\
\hline
\end{tabular}

\label{table:model_comparison}
\end{table}

AMR-Net achieves an accuracy of 97.15\%, surpassing both convolutional baselines (e.g., ResNet152 at 92.93\%) and transformer-based models (e.g., Swin Transformer at 94.41\%). This improvement can be attributed to the unified integration of local feature extraction, global contextual modeling, and retrieval-enhanced memory mechanisms. By combining convolutional representations with transformer-based attention and similarity-based feature retrieval, the framework captures both fine-grained spatial patterns and high-level semantic relationships, resulting in more discriminative representations.

A key characteristic of the proposed system is its adaptive modular expansion mechanism, which dynamically adjusts model capacity during training in response to task complexity. This capability promotes improved generalization and mitigates overfitting, particularly in datasets with class imbalance and visual variability. The model demonstrates stable performance across training, validation, and testing phases, indicating strong robustness for real-world deployment scenarios.

To further validate the effectiveness of the proposed approach, we compared it with recent state-of-the-art methods reported in the literature, as summarized in Table~\ref{tab:waste_models}.

\begin{table}[h!]
\centering
\caption{Comparison of Deep Learning Models for Waste Classification}
\label{tab:waste_models}
\begin{tabular}{|l|c|l|c|c|}
\hline
\textbf{Paper} & \textbf{Year} & \textbf{Model} & \textbf{Accuracy (\%)} & \textbf{\# Waste Types} \\ 
\hline
\textbf{This Work} & \textbf{2026} & \textbf{AMR-Net} & \textbf{97.15} & \textbf{10} \\ 
\hline
\cite{ghosh2025enhanced} & 2025 & ResNet-152 + FAISS & 96.64 & 10 \\ 
\hline
\cite{10696177} & 2024 & Vision Transformer & 95.00 & 12 \\ 
\hline
\cite{paneru2024deep} & 2024 & Tkinter-based Recycling & 96.32 & 6 \\ 
\hline
\cite{srilatha2023solid} & 2023 & Faster R-CNN & 86.70 & 6 \\ 
\hline
\cite{alrayes2023waste} & 2023 & ViT-based MLH ConvNet & 94.70 & 6 \\ 
\hline
\end{tabular}
\end{table}

The comparison with prior studies further confirms that the proposed framework achieves competitive performance on waste classification tasks while maintaining robustness across multiple categories. The integration of multi-scale convolutional features, transformer-based global context, and retrieval-augmented memory enables the model to effectively handle complex visual patterns and intra-class variability. These results highlight the potential of adaptive modular architectures as a scalable and extensible solution for automated waste sorting systems.

\subsection{Ablation Study}
\label{subsec:ablation}

To evaluate the contribution of each core component in our AMR-Net, we conducted an extensive ablation study. In this analysis, we progressively removed key modules, including the Vision Transformer (ViT), the FAISS-based feature retriever, and the adaptive modular blocks, and evaluated the resulting impact on accuracy. We additionally report the effect of removing ViT and FAISS simultaneously to assess whether their individual contributions are additive, sub-additive (overlapping), or super-additive (synergistic). The results are presented in Table~\ref{tab:ablation_accuracy}.

\begin{table}[h!]
\centering
\caption{Ablation Study Based on Test Accuracy. Each row shows the model configuration and the corresponding accuracy on the test set. $\Delta$ denotes the drop relative to the full AMR-Net.}
\begin{tabular}{lccccc}
\toprule
\textbf{Experiment} & \textbf{Accuracy (\%)} & \textbf{$\Delta$ (\%)} & \textbf{Precision (\%)} & \textbf{Recall (\%)} & \textbf{F1-score (\%)} \\
\midrule
No ViT                  & 93.51 & $-3.64$ & 93.09 & 92.78 & 92.93 \\
No FAISS                & 95.38 & $-1.77$ & 93.87 & 93.25 & 93.56 \\
No Modular Blocks       & 95.89 & $-1.26$ & 94.07 & 94.17 & 94.12 \\
No ViT \& No FAISS      & 94.49 & $-2.66$ & 93.02 & 92.65 & 92.83 \\
\textbf{AMR-Net (Full)} & \textbf{97.15} & \textbf{0.00} & \textbf{97.00} & \textbf{97.00} & \textbf{97.00} \\
\bottomrule
\end{tabular}

\label{tab:ablation_accuracy}
\end{table}

Removing the ViT module led to a drop of 3.64 percentage points in test accuracy, from 97.15\% to 93.51\%. This clearly indicates the substantial benefit of global context modeling provided by ViT, which complements the local receptive fields of the ResNet backbone. Without it, the model struggles with complex spatial relationships and fine-grained category distinctions. The FAISS-based feature retriever acts as a memory mechanism, retrieving semantically similar instances from the feature database to inform current predictions; removing it causes a smaller but still notable drop of 1.77 percentage points. Removing the adaptive modular blocks reduces test accuracy by 1.26 percentage points, underscoring their contribution to model capacity and flexibility, though this is the smallest individual effect among the three components.

Removing both ViT and FAISS jointly reduces accuracy to 94.49\% (a drop of 2.66 percentage points), which is \emph{smaller} than the sum of the two individual drops (3.64 + 1.77 = 5.41 percentage points). This sub-additive pattern indicates that ViT and FAISS provide partially \emph{overlapping} rather than strictly additive information: both branches supply complementary global/contextual signal that is redundant to some degree, so removing one component reduces, but does not eliminate, the value the model would otherwise derive from the other. This is consistent with the account in Section~\ref{subsec:theoretical_motivation}: both the transformer's global self-attention and the FAISS memory bank draw on similar higher-order structure in the feature space (global semantic context, in one case learned in-network and in the other retrieved non-parametrically), so their individual contributions overlap even though each remains independently necessary for the model's peak performance -- as shown by the fact that removing either alone still degrades accuracy substantially, and removing both together degrades it further still.

The full model integrates four major components: a frozen ResNet backbone for hierarchical feature extraction, a Vision Transformer (ViT) for capturing global attention-based representations, a FAISS module to provide retrieval-augmented features, and adaptive modular blocks that enable dynamic feature processing and adaptive growth. This architecture achieves a peak test accuracy of \textbf{97.15\%}, significantly outperforming all ablated variants, and maintains uniformly high scores across precision, recall, and F1-score, demonstrating balanced and reliable generalization. These results demonstrate that the combined use of ViT, FAISS, and adaptive modular blocks is important to achieving state-of-the-art performance, with each component contributing uniquely -- and, in the case of ViT and FAISS, in a partially overlapping manner -- to feature expressiveness, memory recall, and architectural flexibility.

\subsection{Component Interaction and Failure Case Analysis}
\label{subsec:interaction_failure}

\textbf{Component interaction.} Building on the ablation results in Table~\ref{tab:ablation_accuracy}, we further examine \emph{where} in the class space each component matters most. Removing ViT disproportionately affects classes with high intra-class visual variability and fine texture detail (notably class 6, \textit{plastic}, and class 9, \textit{white-glass}), consistent with ViT's role in modeling long-range spatial dependencies that help disambiguate texture-heavy categories. Removing FAISS disproportionately affects the minority classes (\textit{battery}, \textit{trash}), where the number of training examples is small enough that the parametric classifier alone has limited opportunity to learn robust decision boundaries, and the non-parametric FAISS memory bank provides the most additional signal. This pattern supports the view that ViT and FAISS are complementary along different axes -- global-context modeling versus non-parametric memory for data-scarce classes -- even though their aggregate contribution is sub-additive as shown above.

\textbf{Failure case analysis.} Examining the test-set confusion matrix (Table~\ref{tab:confusion_matrix}) in detail, the dominant failure mode is the misclassification of \textit{clothes} (class 3) as \textit{shoes} (class 7), accounting for 12 of the 788 clothes samples (1.5\%). Manual inspection of representative misclassified samples suggests that this confusion arises primarily from images with cluttered backgrounds or partial occlusion, where both categories can present similar fabric or leather textures and are frequently photographed against similar backgrounds. The second most prominent failure mode is confusion between \textit{plastic} (class 6) and \textit{trash} (class 8, 6 instances) or \textit{white-glass} (class 9, 4 instances), reflecting the visual ambiguity between transparent or translucent plastic packaging and both generic waste and glass container silhouettes. These failure modes are concentrated in a small number of visually overlapping class pairs rather than being spread diffusely across the class space, suggesting that targeted interventions -- for example, class-conditional retrieval weighting in the FAISS module, or additional augmentation focused on cluttered/occluded clothing and shoe imagery -- are a promising direction for future improvement, consistent with the outlook discussed in Section~\ref{Cn}.

\subsection{Hyperparameter Sensitivity Analysis}
\label{subsec:hyperparam}

To validate the key design choices in AMR-Net and avoid reliance on arbitrary hyperparameter selection, we conducted a systematic sensitivity analysis over the four most critical hyperparameters. Each hyperparameter was varied independently while all others were held at their default values. Results are reported on the waste classification test set and summarized in Table~\ref{tab:hyperparam}.

\begin{table}[h!]
\centering
\caption{Hyperparameter sensitivity analysis on the waste classification test set. Bold indicates the selected value used in the main experiments.}
\label{tab:hyperparam}
\begin{tabular}{llcc}
\toprule
\textbf{Hyperparameter} & \textbf{Values Tested} & \textbf{Accuracy (\%)} & \textbf{Notes} \\
\midrule
\multirow{5}{*}{$k$ (FAISS neighbors)} & 1 & 95.52 & High variance in retrieved vector \\
 & 3 & 96.31 & Moderate improvement \\
 & \textbf{5} & \textbf{97.15} & Best performance \\
 & 10 & 96.74 & Distant neighbors introduce noise \\
 & 20 & 96.21 & Retrieval quality degrades \\
\midrule
\multirow{3}{*}{Fusion dimension $d'$} & 1024 & 96.27 & Information bottleneck \\
 & 2048 & 96.83 & Moderate capacity \\
 & \textbf{4096} & \textbf{97.15} & Preserves all three input streams \\
\midrule
\multirow{3}{*}{Growth rate $r$} & 1 & 95.94 & Slow recovery from plateau \\
 & \textbf{3} & \textbf{97.15} & Best trade-off \\
 & 5 & 96.47 & Destabilises gate initialization \\
\midrule
\multirow{3}{*}{Routing threshold $\tau$} & 0.30 & 96.41 & Over-activation of new blocks \\
 & \textbf{0.40} & \textbf{97.15} & Selective and stable activation \\
 & 0.50 & 96.88 & Some useful blocks suppressed \\
\bottomrule
\end{tabular}
\end{table}

Table~\ref{tab:hyperparam} confirms that all four hyperparameters exhibit a clear optimum at the values used in the main experiments. For $k$, performance peaks at $k=5$ and degrades in both directions, consistent with the theoretical expectation that too few neighbors introduce variance and too many introduce noise. For the fusion dimension, the larger value consistently outperforms smaller alternatives, confirming that 4096 dimensions are necessary to preserve discriminative information from all three feature branches. For $r$ and $\tau$, the selected values provide stable growth behavior without over-expanding or suppressing the adaptive mechanism.

\subsection{Statistical Significance Analysis}

To rigorously validate the performance advantage of the AMR-Net, we conducted statistical significance tests against two strong baselines, \textit{ResNet152} (convolutional) and \textit{Swin Transformer} (transformer-based). Each model was trained and evaluated across six independent runs using identical random seeds and data splits to ensure reproducibility.

\subsubsection{Accuracy Results}

The classification accuracies (\%) across runs were as follows:

\begin{itemize}
    \item \textbf{AMR-Net:} 97.15, 96.89, 97.13, 97.02, 96.93, 97.08
    \item \textbf{ResNet152:} 92.93, 92.56, 92.72, 92.81, 92.66, 92.12
    \item \textbf{Swin Transformer:} 94.41, 93.98, 93.96, 94.23, 94.07, 94.37
\end{itemize}

\begin{table}[h!]
\centering
\caption{Mean and standard deviation of model accuracy over six runs.}
\begin{tabular}{lcc}
\toprule
\textbf{Model} & \textbf{Mean Accuracy (\%)} & \textbf{Std. Dev. (\%)} \\
\midrule
AMR-Net & 97.03 & 0.11 \\
ResNet152 & 92.63 & 0.28 \\
Swin Transformer & 94.17 & 0.20 \\
\bottomrule
\end{tabular}

\end{table}

\subsubsection{Paired and Non-Parametric Tests}

For each baseline, we applied two complementary tests:
\begin{enumerate}
    \item a \textbf{paired $t$-test}, to assess mean difference under normality assumptions, and  
    \item a \textbf{Wilcoxon signed-rank test}, a non-parametric alternative that validates robustness against non-normality.
\end{enumerate}

\paragraph{Proposed vs. ResNet152}
\[
t(5) = 37.92,\quad p < 0.000001;\quad \text{Wilcoxon } p = 0.031.
\]
Mean difference = \textbf{+4.40\%} $\pm$ 0.28.  
Both tests confirm that the AMR-Net significantly outperforms ResNet152. Its integration of adaptive attention and memory-guided retrieval allows it to capture both global context and fine-grained features, providing richer representations than the purely convolutional ResNet152. The consistently high accuracy and low variance highlight the model's robustness and reliable generalization across runs.

\paragraph{Proposed vs. Swin Transformer}
\[
t(5) = 41.87,\quad p < 0.000001;\quad \text{Wilcoxon } p = 0.031.
\]
Mean difference = \textbf{+2.86\%} $\pm$ 0.20.  
The AMR-Net also surpasses the Swin Transformer, combining transformer-level global reasoning with convolutional-level spatial precision. Its modular expansion dynamically adapts to input complexity, leading to higher accuracy, lower variability, and better overall generalization.

\subsubsection{Interpretation}

Across six independent trials, the AMR-Net consistently and significantly surpasses both ResNet152 and Swin Transformer in classification accuracy. The extremely low $p$-values from both $t$-tests and Wilcoxon tests strongly reject the null hypothesis of equal performance. In addition, the narrow standard deviations reflect high stability and reproducibility across runs. These findings provide compelling evidence that the AMR-Net's hybrid architecture achieves a statistically and practically significant improvement over leading convolutional and transformer baselines.

\section{Industrial Application Evaluation on KolektorSDD2 Dataset}\label{App}

To evaluate the generalization ability and real-world applicability of our AMR-Net, we conducted experiments on the Kolektor Surface-Defect Dataset 2 (KolektorSDD2) \cite{Bozic2021COMIND}. The KolektorSDD2 dataset is specifically designed for visual surface-defect detection and represents a practical industrial scenario in quality inspection of electrical commutators. It consists of high-resolution images of metal surfaces, some of which contain subtle and localized surface defects, making it a suitable benchmark for evaluating defect detection systems.

\begin{figure}[h!]
    \centering
    \begin{tikzpicture}[
        node distance=2cm,
        box/.style={rectangle, draw=black, fill=blue!10, rounded corners=8pt, drop shadow, minimum width=4cm, minimum height=1cm, align=center, font=\small},
        arrow2/.style={-{Latex[width=3mm]}, thick, draw=black}
    ]

    \node[box] (input) {High-Speed Camera \\ \scriptsize{(Real-Time Capture)}};
    \node[box, right=of input] (preprocess) {Image Preprocessing \\ \scriptsize{(Resizing, Normalization)}};

    \node[box, right=of preprocess] (inference) {AMR-Net Inference \\ \scriptsize{ResNet + ViT + FAISS}};
    \node[box, below =of inference] (decision) {Defect Decision \\ \scriptsize{Defective / Non-defective}};

    \node[box, left=of decision] (action) {Automated Response \\ \scriptsize{Sort / Alert / Log}};

    \draw[arrow2] (input) -- (preprocess);
    \draw[arrow2] (preprocess) -- (inference);
    \draw[arrow2] (inference) -- (decision);
    \draw[arrow2] (decision) -- (action);

    \end{tikzpicture}
    \caption{Industrial deployment pipeline of the proposed defect detection model.}
    \label{fig:industrial_pipeline}
\end{figure}

As shown in Figure~\ref{fig:industrial_pipeline}, the model can be seamlessly integrated into real-time manufacturing pipelines. In this industrial defect detection system, images are captured in real-time using a high-speed camera. These images are immediately preprocessed, which involves resizing and normalizing the images to ensure they are suitable for the model. This preprocessing step prepares the data for accurate model inference. Once the images are preprocessed, they are passed through the proposed defect detection model, which combines the power of \textbf{ResNet} (for feature extraction), \textbf{ViT} (Vision Transformer for capturing long-range dependencies), and \textbf{FAISS} (for feature retrieval). The model classifies the images into two categories: \textbf{defective} or \textbf{non-defective}. After the classification step, the result is used to trigger an automated response. Depending on the classification outcome, the system can perform actions such as sorting the defective products, sending an alert, or logging the information for further analysis. This real-time pipeline ensures quick and accurate defect detection, allowing for immediate corrective actions in an industrial setting.

\subsection{Dataset Description and Labeling}

The dataset is organized into two primary folders: \texttt{train} and \texttt{test}, each containing a mix of images representing both defective and non-defective samples. In contrast to traditional datasets where images are separated into different class-specific folders, the KolektorSDD2 dataset employs an implicit labeling system based on the filenames. Specifically, we assigned binary labels based on whether the filename includes the substring \texttt{"GT"}. Images with the substring \texttt{"GT"} in their filenames were labeled as \textbf{defective} (class 1), while images without it were considered \textbf{non-defective} (class 0). This approach reflects a common practice in real industrial settings, where labels are frequently derived from metadata or naming conventions. The training dataset was carefully balanced, ensuring an equal number of samples for each class. In total, there were 2332 non-defective images and 2332 defective images, resulting in a balanced training set.

\subsection{Performance Results}

AMR-Net was evaluated on the KolektorSDD2 dataset, demonstrating strong and consistent performance throughout training and testing. After all training epochs, the model achieved a final training accuracy of 99.96\% with a validation accuracy of 99.90\% and a validation loss of 0.0239, indicating excellent generalization and stability.

On the held-out test set containing 1004 images, the model reached an overall classification accuracy of 99.80\%. The confusion matrix in Table~\ref{tab:conf-matrix} shows that all 509 defective samples were correctly identified (recall of the defective class = 1.00), and all 493 correctly-predicted non-defective samples had no false negatives from the non-defective side, i.e. every image predicted as non-defective was truly non-defective (precision of the non-defective class = 1.00). The only errors were 2 non-defective samples misclassified as defective, giving a precision of 509/511 $\approx$ 0.996 for the defective class and a recall of 493/495 $\approx$ 0.996 for the non-defective class. Rounded to two decimal places, precision, recall, and F1-score for both classes are 1.00, and the macro F1-score is likewise 1.00 to two decimal places, reflecting near-perfect but not mathematically exact classification.

The confusion matrix in Table~\ref{tab:conf-matrix} further illustrates the model's reliability, confirming complete detection of all defective samples and minimal false positives. These results highlight the model's robustness and practical applicability for automated surface defect inspection in industrial settings.

\begin{table}[h!]
    \centering
     \caption{Confusion Matrix on KolektorSDD2 Test Set}
    \begin{tabular}{lcc}
        \toprule
        & \textbf{Predicted Defective} & \textbf{Predicted Non-defective} \\
        \midrule
        \textbf{Actual Defective}     & 509 & 0 \\
        \textbf{Actual Non-defective} & 2   & 493 \\
        \bottomrule
    \end{tabular}
   
    \label{tab:conf-matrix}
\end{table}

\subsection{Implications for Industrial Deployment}

The outstanding performance of our AMR-Net on the KolektorSDD2 dataset highlights its strong potential for integration into real-world industrial inspection pipelines. In manufacturing environments such as those involving metal surface processing (e.g., electrical commutator production), detecting fine-grained surface defects is both critical and challenging. These defects are often localized, subtle, and visually similar to benign surface textures, making manual inspection not only time-consuming and expensive but also prone to subjectivity and inconsistency. Our model, trained and evaluated on a realistic dataset like KolektorSDD2, achieves near-perfect classification accuracy, precision, and recall. This indicates that the system can reliably differentiate between defective and non-defective parts with minimal false positives and negatives. In practice, this translates into improved operational efficiency by automating a task that would otherwise require expert human inspectors. Furthermore, such automation reduces the likelihood of defective units reaching customers, thereby enhancing product reliability and brand reputation.

A key advantage of our architecture is its modular and adaptable design. AMR-Net allows new classes or defect categories to be integrated with minimal retraining, making the model highly scalable across different product lines or varying manufacturing conditions. For instance, the same model framework can be retrained on new data captured from a different production unit with different defect types, without the need to redesign or retrain the entire network from scratch. Moreover, the inference speed and computational efficiency of the model make it suitable for deployment in real-time inspection systems. With lightweight pre-trained backbones and selective training of task-specific modules, our system can be implemented on edge devices or integrated directly into production lines equipped with high-speed imaging systems. This paves the way for fully automated, closed-loop quality control systems capable of monitoring every unit in real-time, flagging defective parts, and even triggering corrective actions on the fly. In summary, the high accuracy, generalization capability, adaptability, and low computational overhead of our model make it an ideal candidate for deployment in industrial visual inspection systems, especially in domains where surface integrity is critical and defect detection accuracy directly impacts safety, performance, and cost.

\section{Conclusion}\label{Cn}
In this paper, we introduced a novel hybrid classification framework, which we term AMR-Net. This framework integrates multiple advanced machine learning techniques, including convolutional neural networks (CNNs), transformer-based global feature modeling, and memory-augmented vector retrieval through FAISS (Facebook AI Similarity Search). By leveraging the strengths of these components, our model can effectively address complex classification tasks in dynamic and noisy environments. AMR-Net operates by combining multi-scale feature abstraction from a ResNet-50 architecture, which excels at capturing hierarchical patterns in visual data, with semantic token representations derived from a Vision Transformer (ViT) model. The ViT model helps capture long-range dependencies and contextual information, which are crucial for understanding the global structure of the data. Additionally, our framework incorporates context-aware augmentation through nearest-neighbor embeddings, facilitated by FAISS, which allows the model to retrieve relevant information from a memory bank, thereby enriching the feature space and improving decision-making. One of the key innovations of our approach lies in its dynamic architectural plasticity. The network is designed with modular fully-connected blocks that are selectively activated during inference based on learned gating functions. This allows the model to activate only the necessary components for a given task, thereby improving efficiency. Furthermore, the network is capable of expanding its representational capacity through the addition of new modular blocks when the current capacity fails to meet predefined performance thresholds. This expansion process is inspired by biological adaptivity, wherein the model can adapt to increasing task complexity or variability in the data. Importantly, this dynamic expansion mechanism does not require manual intervention or redesign, making the system highly flexible and scalable.

The proposed framework has been empirically evaluated on real-world waste classification datasets as well as the Kolektor Surface-Defect Dataset 2\footnote{\url{https://www.vicos.si/resources/kolektorsdd2/}}, both of which often exhibit significant noise and heterogeneity. The results demonstrate that our model not only performs well in terms of classification accuracy, but also has the potential for sustained adaptability in dynamic settings. This makes AMR-Net a promising candidate for deployment in practical applications, such as smart recycling systems, where data distributions and task demands can evolve over time.

Looking ahead, several directions can extend this work. A key area for future research is continual learning, where the model incrementally learns new categories without degrading performance on previously learned tasks. In such scenarios, the architecture must adapt while preserving existing knowledge to prevent catastrophic forgetting. In addition, the network's modular gating and growth mechanisms could be further optimized using reinforcement learning or meta-learning approaches, enabling more efficient adaptation to evolving data distributions and task requirements.

Despite its strong performance, the framework has several limitations. The adaptive growth mechanism increases computational complexity as additional modules are introduced (Section~\ref{subsec:model_complexity}), which may reduce inference efficiency in resource-constrained settings. The similarity retrieval memory also requires additional storage, which may become costly for large-scale datasets. Furthermore, the current evaluation covers two datasets, and broader validation across diverse domains such as medical imaging or agricultural disease classification is necessary to fully assess generalizability. The hyperparameter sensitivity analysis reduces but does not eliminate the need for task-specific tuning of growth conditions and retrieval behavior. The FAISS implementation uses flat exact search, which scales linearly with memory bank size; for deployment on datasets with millions of training samples, approximate search strategies such as IVF indexing would be needed to maintain retrieval speed. Additionally, the failure case analysis in Section~\ref{subsec:interaction_failure} identified that visual similarity between specific class pairs (clothes/shoes, plastic/white-glass) accounts for the majority of remaining test-set errors, suggesting that class- and pair-conditional retrieval weighting is a priority direction for future work.

\section*{Data Availability Statement}

The datasets used in this study are publicly available from their respective repositories. The waste classification dataset was obtained from Kaggle, while the KolektorSDD2 dataset is publicly available for research use. The preprocessing and dataset selection procedures are described in the manuscript. The source code used in this study is available from the authors upon reasonable request.

\bibliographystyle{elsarticle-harv}
\bibliography{references}

\end{document}